# Smartphone-based Home Robotics

Lojain Jibawi, Saoussen Said, Kenjiro Tadakuma, and Jose Berengueres

*Abstract*— Humanoid robotics is a promising field because the strong human preference to interact with anthropomorphic interfaces. Despite this, humanoid robots are far from reaching main stream adoption and the features available in such robots seem to lag that of the latest smartphones. A fragmented robot ecosystem and low incentives to developers do not help to foster the creation of Robot-Apps either. In contrast, smartphones enjoy high adoption rates and a vibrant app ecosystem (4M apps published). Given this, it seems logical to apply the mobile SW and HW development model to humanoid robots. One way is to use a smartphone to power the robot. Smartphones have been embedded in toys and drones before. However, they have never been used as the main compute unit in a humanoid embodiment. Here, we introduce a novel robot architecture based on smartphones that demonstrates x3 cost reduction and that is compatible with iOS/Android.

## I. INTRODUCTION

Humanoid robots offer undeniable user experience advantages compared to other human-machine interfaces due to the preference that humans have for anthropomorphic communication [1]. Another advantage of humanoids is backwards compatibility with the existing tools and utensils that we already have such as cars, excavators and trays [2]. In addition, anthropomorphism can also be leveraged to improve industrial robots because it reduces miss-communication between the machine the human. For example, Baxter displays a face on an LCD screen. Before Baxter moves an arm, the eyes in the LCD gaze to the future position of the arm. Signaling future motion through gaze lowers the risk of a fortuitous collision with the human operator [3]. Humanoids are also used in the fields of therapeutics (to treat autism) and in social robotics, to name a few. The benefits of using a humanoid interface are multiple. However, why then is humanoid adoption rate still so low?

### A. Commercial humanoids

One explanation is the high cost of the hardware itself. Some hyper-realistic soft-shell robots such as Geminoid [4] have received significant media attention but failed to reach commercialization due to cost. Lower cost hard-shell humanoids exist too. The closest example to a commercial success is perhaps Softbank's Pepper(~$20k) and Nao(~$10k) with approximately 1k and 10k units sold.

*Research supported by UAEU UPAR G25002 grant

L.J and S. S are with VOTEK FZ LCC, Dubai Internet City, UAE (e-mail: lojain@votek.me , sawsan@votek.me ).

K. T. is with the robots lab. At Tohoku University, Sendai, Japan (e-mail: kenjiro.tadakuma@tohoku-u.ac.jp).

J. B., is with CS Dept. at CIT, UAE University, Al Ain, AD, 17551 UAE. (tel: +971553519573, e-mail: jose@uaeu.ac.ae).

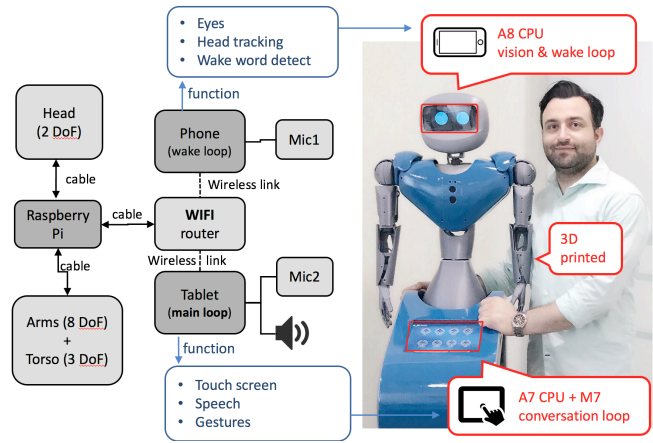

Fig.1 By using mobiles in robots, mobile developers can leverage existing iOS/Android know-how to make powerful Robot-Apps.

However, their utility seems limited to selfie taking, store greeting and news reading. Pepper's abilities are limited by the processing power of an Intel Atom CPU (512k L2 cache memory). There are less than 1000 Apps available in its own Robot-App store [5]. This contrasts with the mass market successes of the iRoomba vacuum cleaner (20M units sold) or the UI-less anthropomorphic-voice interface Amazon Echo (44M sold and 20k skills in the Alexa store). Never the less the commercial success, these (more utilitarian) less anthropomorphic products have not captured the popular imagination in the same way [4, 5] have. The case for humanoid robots seems compelling and the psychological theory behind is also well established [1].

### B. Smartphones

On the other hand, smartphones, (with close to 2.5bn units in use daily), enjoy a global adoption rates close to 50%. Smartphone hardware is also ahead of humanoid hardware in terms cost and performance and has a shorter life cycle than robots. While smartphones have a cycle of 18 months, the Nao robot for instance, has a cycle close to that of automobiles (7 years). When we look at Apps available, the gap is even more dramatic. The largest mobile App store in the world, has 2.4 million apps available [6]. In contrast, the most successful Robot-App store today has less than 1000 Apps, only 10k potential users (corresponding to the total units sold) and, no App monetization mechanism in place [5].

### C. Incentives

However, without incentives there will be no great Apps and without great Apps robots are just empty boxes. Then, how to attract top developer talent to build Apps for robots? One way to incentivize existing developers is to let them use the SDKs they already use to make mobile Apps. In Fig. 1, we show an example. A 3D printed humanoid is powered by smartphones to perform AI functions similar or better than a Softbank's Pepper unit. In the following section, we layout this proposed novel robot architecture based on smartphone hardware. Then we explain one implementation case where an existing mobile App was ported to run on the said architecture (Fig. 1). Finally, we compare the humanoid performance to its closest alternative (Pepper).

## II. REQUIREMENTS

### A. From App to humanoid

The MHRE is a housing center that provides housing to Emirati families of Dubai [7]. The center has a mandate to go paperless by 2020. Therefore, an App was commissioned where users can use and apply for various services that the center provides such as to book an appointment, to look at the status of the loan and to request or fill required forms by the housing program.

### B. Mobile App description

The App uses an API that interfaces with a backend server that is connected to a custom management software (ERP/SAP). In total, about 129 different actions are handled via the API and 259 different types of requests or forms can be logged by customers using the App. However, many of the MHRE users trust paper more than pixels due to age and cultural factors. A previous version of the app is available at [7].

### C. Humanoid requirements

Following the successful development and test of the new mobile App, the center commissioned a companion humanoid robot to the App to help such users to transition to paperless. The humanoid has the same functionality as the App. It usually sits in the lobby of the MHRE center and it is also used in promotional events. Hence, in addition to the mobile App functionally, the humanoid has the following humanoid-specific requirements:

1. Wake word function (*OK Google/ Alexa / Hey Siri-*style). Response time < 200ms. Signal when robot is listening
2. Arm gesture matching robot speech for improved communication [8]
3. Head and tracking of the human faces for eye contact and improved empathy response [9]
4. Gamification elements:
   a. Age estimation game (FACE API, Microsoft Azure)
   b. Take-a-selfie pose
   c. Eye shape changes with context

### D. State-of-the-art

Previously, a few robot applications have used mobile devices as compute units. Romo is a tracked-mobile platform where an iPhone could be docked to give it mobility. Double Robotics is an alternative to Suitabletech's Beam robot, where an iPad can be mounted onto of a self-balancing wheel to serve as teleconference device with mobility. Using smartphones to power edu-toys has also been executed commercially. An example by some of the authors here is [10]. Mobile devices have also been integrated in drones too [11]. However, using exclusively mobile devices to fully power a humanoid robot was never attempted before. Here we combined the processing power of two mobile devices to reach a combined processing power that yields a minimally acceptable user experience; ie. a response time to user interactions below 200ms and, speech recognition error rates similar to best in class systems (5%).

### E. Main compute unit (Tablet)

We have used an Apple iPad Air from the year 2013 as the main compute unit of the robot. The tablet runs an iOS app that controls the conversation loop. This loop is in charge of fulfilling all the same functionality of the mobile App (log in, filling forms, submit requests) by using voice as input method. It uses the same VOTEK Arabic speech recognition libraries used in [10], Arabic Dubai dialect, 15k words. The app is connected to the backend of the center via the same API that the stand alone mobile App uses. Hence, through this API the robot can submit forms that have been filled by speech recognition using the robot as interface.

### F. Secondary compute unit (iPhone)

The tablet's A7 processor was not powerful enough to accomplish all the tasks without appearing laggy to the user. Therefore, a second mobile device was added: an iPhone 6 plus. There were four reasons to add a second compute unit at the expense of increased complexity (two Apps instead one) and cost (+$1k). The critical one was to perform wake word detection efficiently. We were not able to accomplish this by using the iPad alone. A secondary function was to serve as a high-quality display to show the eyes of the humanoid (the most looked-at part of a humanoid). The eyes are animated to enhance emotion expression and other useful cues such as: *wait, I am processing something*. The wake word is *ya-Rashid*, which translates to English as *Hey, Mr. Rashid*. The speech loop and the wake word loop do not use cloud speech APIs. This enhances user privacy, response times and security against hacking. A third function is head tracking; the iPhone 6 plus acclaimed camera can be used to track the user face and for eye contact. Finally, a wealth of existing know-how, libraries and Apps **native** to the operating system of the smartphone and tablet exist already. This makes it straightforward to port or integrate them to work in the humanoid, something not possible with Pepper. However, we used one platform agnostic cloud service, (Microsoft Azure FACE API [12]), to include a small gamification element: The users can take a photo and the robot guesses their age.

### G. Detailed justification of using two compute units

As mentioned, the main reason to use two mobile devices instead of just a single device is that when the wake word loop runs in the same device as the conversation loop the response time increases beyond 300ms in our tests. This resulted in a poor user experience as the robot appears laggy. By using a second device dedicated to wake word detection we achieve an average wake word detection response time (from the word ending) of 105ms (similar or better than Alexa, Siri and Google home). Running two separated

Automatic Speech Recognition (ASR) systems, one for Wake word spotting (WWS) and the other for the LVCSR (large vocabulary continuous speech recognition) has also the following benefits:

(i) We avoid continuous context switching between WWS mode and main recognition mode during conversation.

(ii) We make it more accurate for a user to interrupt a conversation (dialog loop interrupt).

This set up, lets the LVCSR to focus in its main task without loss in performance and with limited processing resources.

### H. Wake word detection system design considerations

The wake word detection system (aka, keyword spotting system) can be considered as a special case of the standard Speech to Text engine in which the vocabulary is restricted to the keywords vs. non-keywords. In order to perform an efficient keyword triggering with low error rate several approaches were tested. A good technical explanation was given by [13]. Here are three we tested:

*1) Phonetic Search*

This is based on a phoneme decoder which transforms speech input into a phoneme textual sequence using phoneme transition probabilities. Then, a distance measured between target keyword and the resulting phonemes sequence is computed. This method results in a high false positive rate (5.5%).

*2) Modified hidden Markov model approach*

A modified mechanism to the ordinary connected word recognition based on HMM also has been tried in order to recognize a predefined keyword in an unconstrained / unconnected speech style. It depends on applying multiple statistical models representing the actual vocabulary and background speech. This method performance is superior than phonetic search but false positive rate where rate still high (1-2%).

*3) Language model approach*

We apply a full speech recognition on the entire stream of speech using a Markov Model like in a full Automatic Speech Recognition (ASR) system but with small set of vocabulary domain (a 3-gram language model). With an improved scoring technique, this method yielded the lowest false positive rate < 0.01.

### I. Gestures

The iPad and the iPhone are wirelessly connected to a of-the-shelf Wi-Fi router, which is connected to the internet and connected by cable to a Raspberry Pi. The Raspberry pi has been prepared to use Poppy robots software (the Open source robotic platform) [14]. 13 servomotors are controlled in this way to move the arms, torso and head. They are linked by cables in series. The servomotors are used to realize ten prerecorded gestures (rise hand, shake, wave, selfie pose…) and head tracking which is performed by two motors in the neck and by the torso when the neck reaches a limit angle (see video). The gestures of the body and arms are controlled by the conversation loop (in the iPad) the head tracking is controlled independently by the computer vision loop which runs in the iPhone. Fig. 2 shows a rendering with the position of the servomotors. To evaluate the proposed architecture, in the following section we show (i) an implementation that uses the said architecture and (ii) evaluate it to alternative solutions in terms cost and performance. Qualitative user experience notes are given too.

### III. EVALUATION

A humanoid robot "Mr. Rashid" was developed applying the framework proposed. Table 1 shows HW specifications. Table 2 shows performance indicators of the software. Fig. 3 visualizes development costs in terms of source code and compares it to the mobile App. 75,844 SLOCS or lines of standard code were used to make the mobile app. To make the App for the humanoid based on the mobile App (110,315 SLOCS were required (82,279 for the tablet App and 28,036 for the App that runs in the iPhone)

TABLE I. SPECIFICATIONS HW

| Item | data |
| --- | --- |
| Weight | PLA 3D-printed exoskeleton 3.2Kg<br>Wooden base: 7Kg<br>Other components ~1Kg |
| Dimensions | 166 x 55 x 28 (cm) |
| CPU | Main: A7/M7 (iPad)<br>Secondary: A8 (iPhone) |
| Servos | 1 x Dynamixel AX12A (neck base)<br>3 x Dynamixel MX106 (torso base)<br>9 x Dynamixel MX28TA, (72g)<br>1 x DVE Power source (60W)<br>1 x Raspberry Pi 3 (controller) |
| Wirelss Router | Tplink-300nbWR840nnk |

TABLE II. SPECIFICATIONS SW

| Item | data |
| --- | --- |
| Wake word | *ya rashid* response time: 105ms<br>Experimental False Positive rate* < 99.9% (similar to Alexa) |
| Speech recognition Error rate | 6% (VOTEK, 15k Arabic, words, quiet room)<br>(Best ASR 5%) |
| Core functionality | Same as mobile app<br>                          (72 k SLOCS) |
| Extra functions | Age game<br>Selfie game<br>Command gesture from icon menu<br>Head tracking<br>                          (111 k SLOCS) |
| predefined gestures | 10 (Pepper 3) |
| Head tracking | By user face detection. (1s to correct a 6-degree instantaneous change) |
| Eye animations and eye color transition codes | Blue to Green → Wake word detected, listening to Speech (you got my attention)<br>Green to blue → Stopped listening (not paying attention to you)<br>Any color to Red→ Error<br>Clock icon → Wait<br>Map icon → Showing map in the tablet<br>Camera icon→ Taking photo |

\* wake up word falsely detected

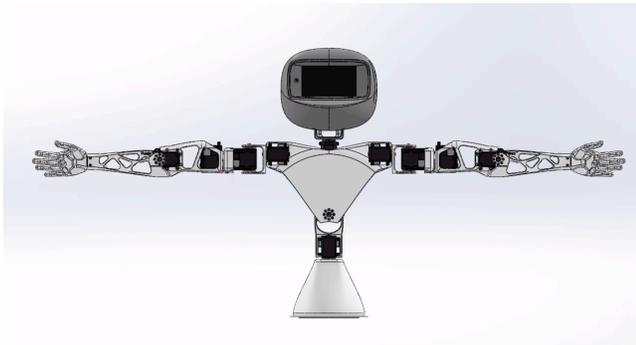

Fig. 2 A 3D rendering showing the parts and the 13 servos that are used. Arms (8), head (2) and torso (3). Using 3D printing and consumer grade mobile devices lowers HW costs by a factor of 3 as compared to the Pepper robot,

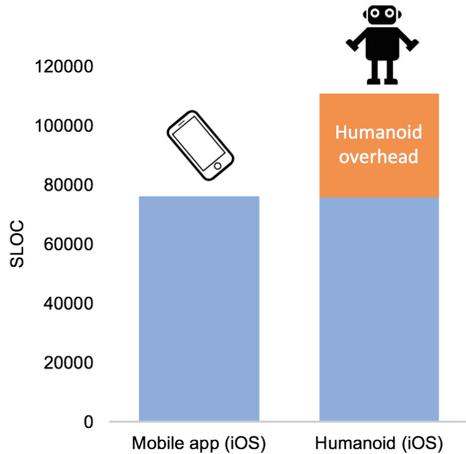

Fig. 3 Building a humanoid version of the mobile App only took an additional 31% SLOCs. 90% of the mobile App code was reused.

### A. From mobile App to humanoid App

To develop the initial stand-alone mobile app took 6 man x months and 75k lines of source code not including external libraries. This includes 4 complete UI redesigns. However, porting the mobile app to the humanoid increased the lines of code by 31% and took only 2 extra man x months with the same developers (senior developers). 90% of the existing code was reused. Fig. 4 shows the flow between the wake word loop and the conversation loop.

### B. User experience notes

The robot was tested for 2 months at the MHRE center lobby and for one week at the Festival city shopping mall, Dubai. From a qualitative point of view, an overwhelming majority of users were more impressed with visual recognition functions of the robot (age guess game) than with speech recognition abilities, even though the latter was costlier to the development team than the former (FACE API). In user tests, workers in the center realized that users feel more **comfortable** interacting with the robot than with a human because they feel the robot does not judge them if they asked silly questions. A reason in agreement with why teenagers prefer to learn via Khan academy videos rather than being lectured in person by a human teacher [15].

### C. Comparison to Pepper

Table 4 compares Mr. Rashid humanoid robot to the closest competing alternatives

TABLE III. COMPARISON

| metric | iOS Mobile app | Mobile based Humanoid (Mr. Rashid) | Softbank Pepper |
|---|---|---|---|
| Year model | 2017 | 2018 | 2017 |
| CPU (L2+L3 cache) | - | A8 + A7 5 + 5MB | Atom Z350 512 kB |
| Autonomy | 8h (typ.) | Plug-in | 12h (typ.) |
| **Wake word** | - | Yes | - |
| **Head tracking** | n/a | Yes | Yes |
| DoF | 0 | 13 | 20 |
| **Camera** model | **8 MP** iPhone6+ | **8 MP** iPhone6+ | **5 MP** OV5640 |
| Platform (developers) | Apple iOS / Android **(12M)** | | NAOqi OS (~1k) |
| Potential user base | 2.5 bn (smartphones) | 1 | 10k (units sold) |
| Number of Apps available | 2.8M/2.2M (iOS/Android) | 1 | ~1k |
| App dev. cost (man xmonth) | 6 | 6[a] + 2 | ? |
| HW cost | $0.3 - 0.9 | ~$6.8k | ~$20k |
| **HW cost /DoF** | - | **~$500** | **~$1000** |

a. 90% of code from App reused.

### IV. CONCLUSIONS

We have shown how build a fully functioning humanoid robot based on mobile devices. By doing that, resources, that are commonly available to largest and most talented pool of software developers (mobile App developers) can also be applied to develop Robot-Apps.

The robot developed has hardware specifications similar or better to a Pepper robot (except mobility) yet costs about 1/3rd. More importantly, application development cycle is quicker (and cheaper) because the mobile iOS development ecosystem is several orders of magnitude larger for iOS (and Android) than for Pepper's NaoQi. SDK, community, support, know-how, libraries available, APIs available, and number of developers available for projects are also more favorable to mobile platforms than any of the existing robotic platforms.

Finally, because the robot only uses consumer grade devices as compute units, the robot is as easy to upgrade (or repair) as replacing the tablet or the phone with a new model. This is currently not feasible in many alternative robots such as Pepper as they are based on non-standard mechanical and electronic parts that are located in hard to access places.

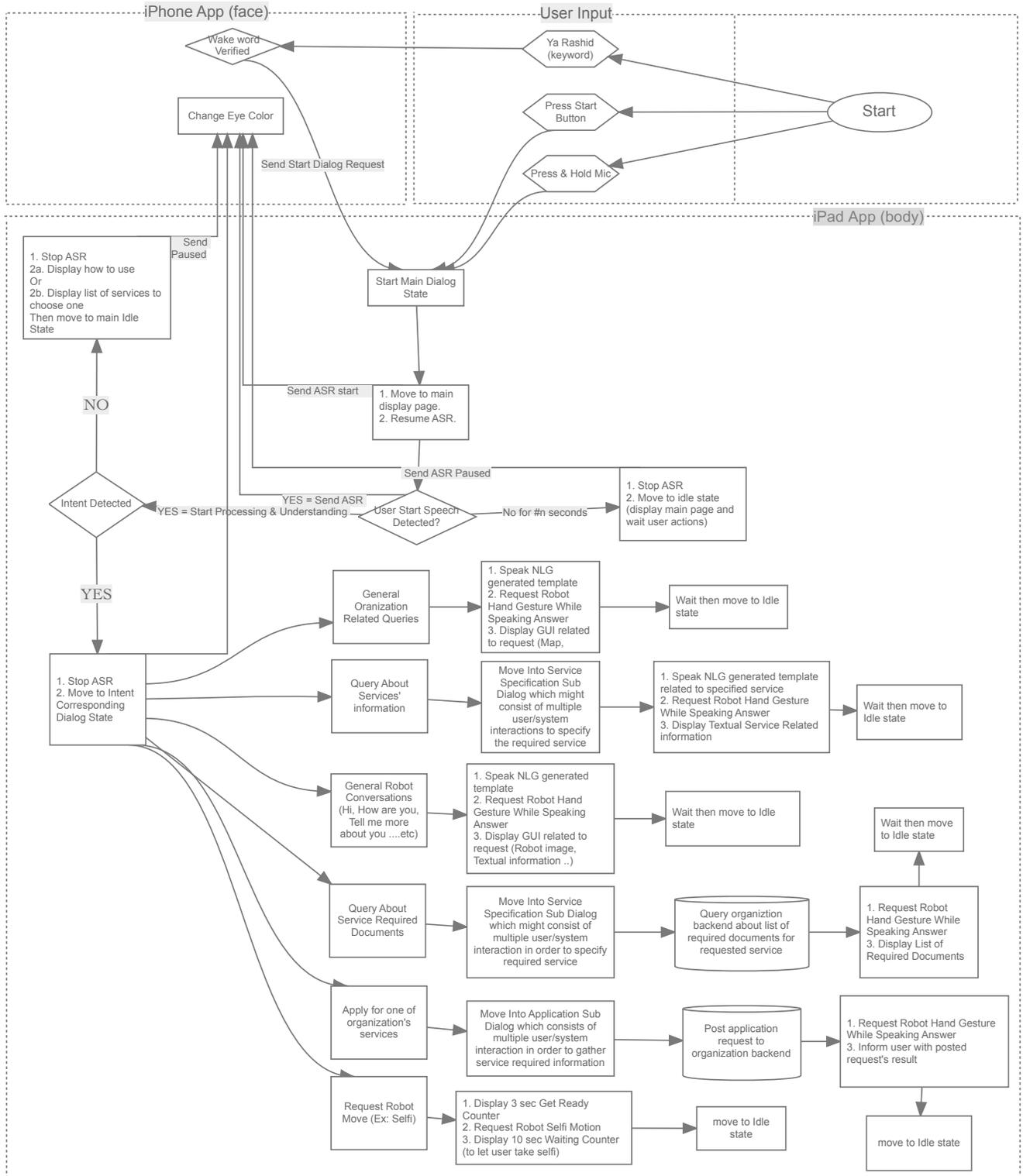

Fig. 4 Master Interaction flowchart. The iPhone performs the wake word detection loop. The iPad processor performs the Main conversation loop, and controls the gestures, API with backend, and eye color (iPhone). Computer vision functions performed in the iPhone are not shown for simplicity. Separating wake word from the main loop improves response time with limited hardware.

### A. Towards a humanoid app store

As we saw in section 4.B, converting an existing mobile app to humanoid app i.e., *humanizing* the app, results in better user experience at a relative 30% increase in lines of source code which is a proxy for cost. Since most consumer facing software being developed today is mobile first, it seems logical that the first humanoid apps for domestic robots will be based on existing mobile apps. This, combined with the fact that one in two software developers in the world today are mobile software developers, is a strong argument in favor of using mobile platforms for humanoid robots.

To conclude, a critical element to having a thriving App ecosystem is the incentive scheme (currently lacking in Robot-App stores). In contrast, both the Android store and the Apple have the trust of customers, their credit card numbers and a robust payment system. If robot makers want to have more Apps in their platforms, making them compatible with mobile operating systems seems a logical step towards attracting top talent. We hope this case study helps ease the development of humanoid Robot-Apps.

ACKNOWLEDGMENT

We would like to thank Mohammed Bin Rashid Housing Establishment employees for its cooperation with usability testing and user insights.

REFERENCES

[1] Epley N, Waytz A, Cacioppo JT. On seeing human: a three-factor theory of anthropomorphism. Psychological review. 2007 Oct;114(4):864.
[2] Thrun S, Montemerlo M, Dahlkamp H, Stavens D, Aron A, Diebel J, Fong P, Gale J, Halpenny M, Hoffmann G, Lau K. Stanley: The robot that won the DARPA Grand Challenge. Journal of field Robotics. 2006 Sep 1;23(9):661-92.
[3] Guizzo E, Ackerman E. How rethink robotics built its new baxter robot worker. IEEE spectrum. 2012 Sep 18:18.
[4] Nishio S, Ishiguro H, Hagita N. Geminoid: Teleoperated android of an existing person. InHumanoid robots: new developments 2007. InTech.
[5] https://store.aldebaran.com/
[6] Martin W, Sarro F, Jia Y, Zhang Y, Harman M. A survey of app store analysis for software engineering. IEEE transactions on software engineering. 2017 Sep 1;43(9):817-47.
[7] https://new.mrhe.gov.ae/en/eServices/SmartServices/Pages/default.aspx
[8] Stiefelhagen R, Fugen C, Gieselmann R, Holzapfel H, Nickel K, Waibel A. Natural human-robot interaction using speech, head pose and gestures. InIntelligent Robots and Systems, 2004.(IROS 2004). Proceedings. 2004 IEEE/RSJ International Conference on 2004 Sep (Vol. 3, pp. 2422-2427). IEEE.
[9] Breazeal C, Kidd CD, Thomaz AL, Hoffman G, Berlin M. Effects of nonverbal communication on efficiency and robustness in human-robot teamwork. InIntelligent Robots and Systems, 2005.(IROS 2005). 2005 IEEE/RSJ International Conference on 2005 Aug 2 (pp. 708-713). IEEE.
[10] http://loujee.com/en/#video
[11] Loianno G, Mulgaonkar Y, Brunner C, Ahuja D, Ramanandan A, Chari M, Diaz S, Kumar V. Smartphones power flying robots. InIntelligent Robots and Systems (IROS), 2015 IEEE/RSJ International Conference on 2015 Sep 28 (pp. 1256-1263). IEEE.
[12] Del Sole A. Introducing Microsoft Cognitive Services. In Microsoft Computer Vision APIs Distilled 2018 (pp. 1-4). Apress, Berkeley, CA.
[13] Michaely AH, Zhang X, Simko G, Parada C, Aleksic P. Keyword Spotting for Google Assistant Using Contextual Speech Recognition. InProceedings of ASRU 2017.
[14] Lapeyre M, Rouanet P, Grizou J, Nguyen S, Depraetre F, Le Falher A, Oudeyer PY. Poppy project: open-source fabrication of 3D printed humanoid robot for science, education and art. In Digital Intelligence 2014 2014 Sep 17 (p. 6).
[15] Thompson C. How Khan Academy is changing the rules of education. Wired Magazine. 2011 Jul 15;126:1-5.